\documentclass[11pt,a4paper]{article}
\usepackage[hyperref]{naaclhlt2019} 
\usepackage{caption}
\usepackage{subcaption}
\usepackage{times}
\usepackage{latexsym}
\usepackage{amsmath}
\usepackage{tipa}
\usepackage{booktabs}
\usepackage{pifont}
\usepackage{url}
\usepackage{amsmath}
\usepackage{amssymb}
\usepackage{multirow}
\usepackage{xspace}
\usepackage{soul}
\usepackage{subcaption}

\newcommand{\cmark}{\ding{51}}%
\newcommand{\xmark}{\ding{55}}%
\newcommand{\ra}[1]{\renewcommand{\arraystretch}{#1}}

\newcommand{\softmax}{\mathrm{softmax}}

\newcommand*{\img}[1]{%
    \raisebox{-.1\baselineskip}{%
        \includegraphics[
        height=0.8\baselineskip,
        width=0.8\baselineskip,
        keepaspectratio,
        ]{#1}%
    }%
}

\newcommand{\sys}{\img{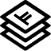}\xspace\text{Multee}\xspace}

\newcommand{\setto}{\overset{\Delta}{=}}

\aclfinalcopy 

\usepackage{color}
\usepackage{booktabs}
\usepackage{graphicx}

\newcommand{\eat}[1]{}

{\list{}{\leftmargin=#1\rightmargin=#1}\item[]}%
{\endlist}

\usepackage{etoolbox}
\makeatletter
\patchcmd\@combinedblfloats{\box\@outputbox}{\unvbox\@outputbox}{}{\errmessage{\noexpand patch failed}}
\makeatother

\title{Repurposing Entailment for Multi-Hop Question Answering Tasks}

\author{
Harsh Trivedi$^\clubsuit$, Heeyoung Kwon$^\clubsuit$, Tushar Khot$^\spadesuit$, Ashish Sabharwal$^\spadesuit$, Niranjan Balasubramanian$^\clubsuit$\\
\hspace{1ex}\\  
$^\clubsuit$ Stony Brook University, Stony Brook, U.S.A.\\
\texttt{\{hjtrivedi,heekwon,niranjan\}@cs.stonybrook.edu}\\
$^\spadesuit$ Allen Institute for Artificial Intelligence, Seattle, U.S.A.\\
\texttt{\{tushark,ashishs\}@allenai.org}
}

\begin{document}

\maketitle

\begin{abstract}
Question Answering (QA) naturally reduces to an entailment problem, namely, verifying whether some text entails the answer to a question. However, for multi-hop QA tasks, which require reasoning with \emph{multiple} sentences, it remains unclear how best to utilize entailment models pre-trained on large scale datasets such as SNLI, which are based on sentence pairs.

We introduce \sys, a general architecture that can effectively use entailment models for multi-hop QA tasks. \sys uses (i) a local module that helps locate important sentences, thereby avoiding distracting information, and (ii) a global module that aggregates information by effectively incorporating importance weights. Importantly, we show that both modules can use entailment functions pre-trained on a large scale NLI datasets. We evaluate performance on MultiRC and OpenBookQA, two multihop QA datasets.  When using an entailment function pre-trained on NLI datasets, \sys outperforms QA models trained only on the target QA datasets and the OpenAI transformer models. The code is available at \url{https://github.com/StonyBrookNLP/multee}.
\end{abstract}

\section{Introduction}

How can we effectively use textual entailment models for question answering? Previous attempts at this have resulted in limited success~\cite{harabagiu2006methods,sacaleanu2008entailment,Clark2012AnEA}. With recent large scale entailment datasets~\citep{bowman2015large,williams2017broad,khot2018scitail} pushing entailment models to high accuracies~\cite{chen2016enhanced,parikh2016decomposable,wang2017bilateral}, we re-visit this challenge and propose a novel method for re-purposing neural entailment models for QA.

\begin{figure}[t!]
    \centering
    \includegraphics[width=0.48\textwidth]{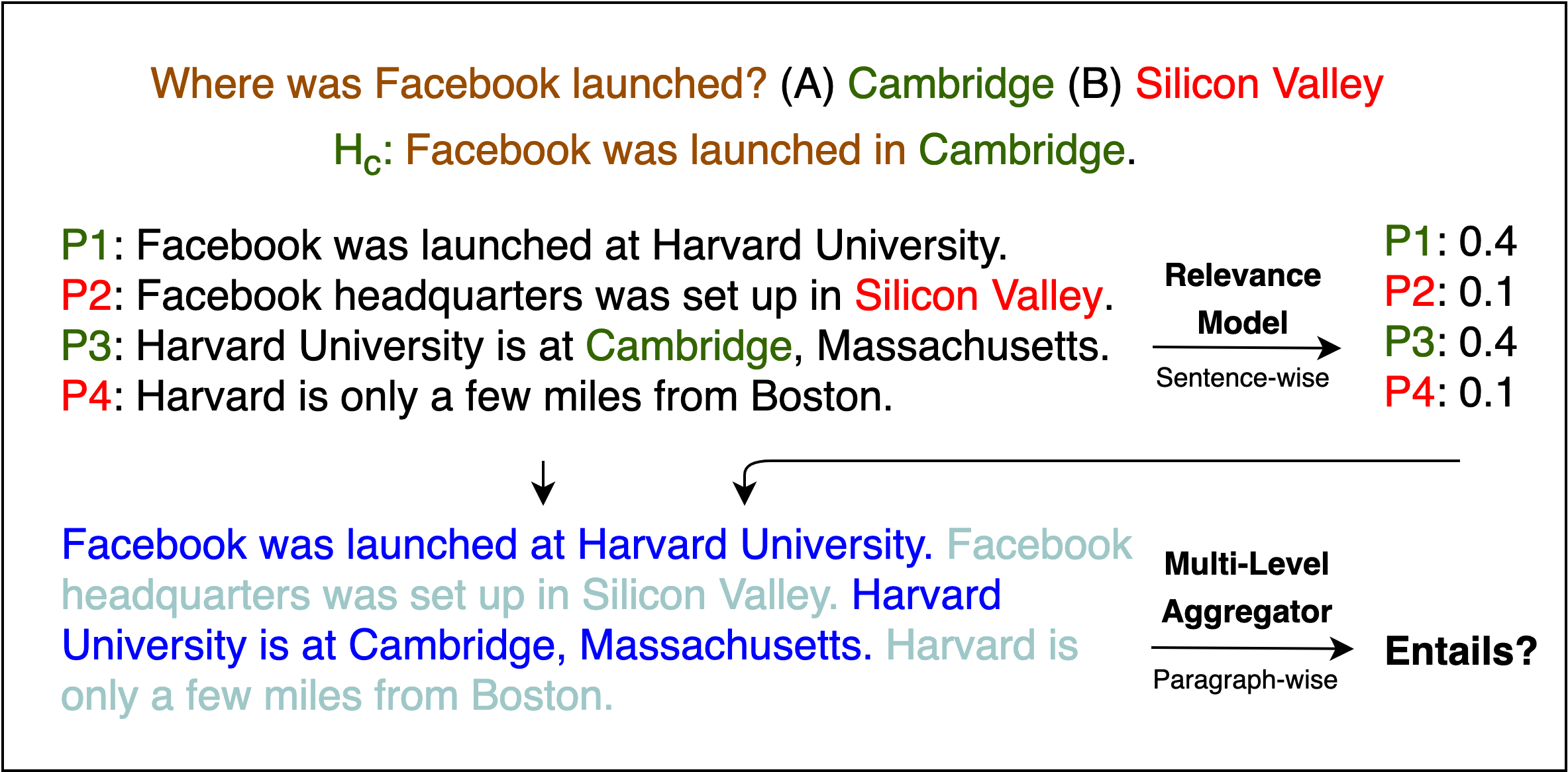}
    \caption{An example illustrating the challenges in using sentence-level entailment model for multi-sentence reasoning needed for QA, and the high-level approach used in \sys.}
    \label{fig:example}
    \vspace{-1em}
\end{figure}

A key difficulty in using entailment models for QA turns out to be the mismatch between the inputs to the two tasks: large-scale entailment datasets are typically framed at a \emph{sentence level}, whereas question answering requires verifying whether \emph{multiple sentences}, taken together as a premise, entail a hypothesis.

There are two straightforward ways to address this mismatch: (1) aggregate independent entailment decisions over each premise sentence, or (2) make a single entailment decision after concatenating all premise sentences. Neither approach is fully satisfactory. To understand why, 
consider the set of premises in Figure~\ref{fig:example}, which entail the hypothesis $H_c$. Specifically, the combined information in $P1$ and $P3$ entails $H_c$, which corresponds to the correct answer \textit{Cambridge}. On one hand, aggregating independent decisions will fail because no individual premise entails $H_C$. On the other hand, simply concatenating premises to form a single paragraph will fail because distracting information in $P2$ and $P4$ can muddle useful information in $P1$ and $P3$. An effective approach, therefore, must recognize relevant sentences (i.e., avoid distracting ones) and compose their sentence-level information.

Our solution to this challenge is based on the observation that a sentence-level entailment function can be re-purposed for both recognizing relevant sentences, and for computing sentence-level representations. Both tasks require comparing information in a pair of texts, but the objectives of the comparison are different. This means we can take an entailment function that is trained for basic entailment (i.e., comparing information in texts), and adapt it to work for both recognizing relevance and computing representations. Thus, this architecture allows us to incorporate advances in entailment architectures and to leverage pre-trained models obtained using large scale entailment datasets.

To this end, we propose a general  architecture that uses a (pre-trained) entailment function $f_e$ for multi-sentence QA. Given a hypothesis statement $H_{qa}$ representing a candidate answer, and the set of premise sentences $\{P_i\}$, our proposed architecture uses the same function $f_e$ for two components: (a) a sentence relevance module that scores each $P_i$ based on its potential relevance to $H_{qa}$, with the goal of weeding out distractors; and (b) a relevance-weighted aggregator that combines entailment information from multiple $P_i$. 

Thus, we build effective entailment aware representations of larger contexts (i.e., multiple sentences) from those of small contexts (i.e., individual sentences). The main strength of our approach is that, unlike standard attention mechanisms, the aggregator module uses the attention scores from the relevance module at \textit{multiple} levels of abstractions (e.g., multiple layers of a neural network) within $f_e$, using \textit{join} operations that compose representations at each level. We refer to this \textbf{mu}lti-\textbf{l}evel aggregation of \textbf{te}xtual \textbf{e}ntailment representations as \sys (pronounced multi).  

Our implementation of \sys uses ESIM~\cite{chen2016enhanced}, a recent sentence-level entailment model, pre-trained on SNLI and MultiNLI datasets. We demonstrate its effectiveness on two challenging multi-sentence reasoning datasets: MultiRC~\cite{MultiRC2018} and OpenBookQA~\cite{mihaylov2018can}. \sys using ELMo contextual embeddings~\cite{Peters:2018} matches state-of-the-art results achieved with large transfomer-based models~\cite{radford2018improving} that were trained on a sequence of large scale tasks~\cite{sun2018improving}. 

Ablation studies demonstrate that both relevance scoring and multi-level aggregation are valuable, and that pre-training on large entailment corpora is particularly helpful for OpenBookQA.

This work makes three main contributions: \textbf{(i)} A novel approach to use pre-trained entailment models for question answering. \textbf{(ii)} A model that incorporates local (sentence level) entailment decisions with global (document level) entailment decisions to effectively aggregate information for multi-hop QA task. \textbf{(iii)} An empirical evaluation that shows entailment based QA can achieve state-of-the-art performance on two challenging multi-hop QA datasets, OpenBookQA and MultiRC.

\section{Question Answering using Entailment}

Non-extractive question answering can be seen as a textual entailment problem, where we verify whether a hypothesis constructed out of a question and a candidate answer is entailed by the knowledge---a collection of sentences\footnote{This collection can be a {\em sequence} in the case of passage comprehension or a {\em list} of sentences, potentially from varied sources, in the case of QA over multiple documents.} in the source text. The probability of an answer $A$, given a question $Q$, can be modeled as the probability of a set of premises $\{P_i\}$ entailing a hypothesis statement $H_{qa}$ constructed from $Q$ and $A$:
\begin{align}
    \Pr[A \mid Q,\{P_i\}] & \ \setto \ \Pr[\{P_i\} \vDash H_{qa}]
\end{align}
Here we use $\vDash$ to denote textual entailment.
Given QA training data, we can then learn a model that approximates the entailment probability $\Pr[\{P_i\} \vDash H_{qa}]$.

Can one build an effective QA model $g_e$ using an existing entailment model $f_e$ that has been pre-trained on a large-scale entailment dataset? Figure~\ref{fig:max-concat} illustrates two straightforward ways of doing so, using $f_e$ as a black-box function:

\begin{figure}[ht]
    \centering
    \includegraphics[width=0.45\textwidth]{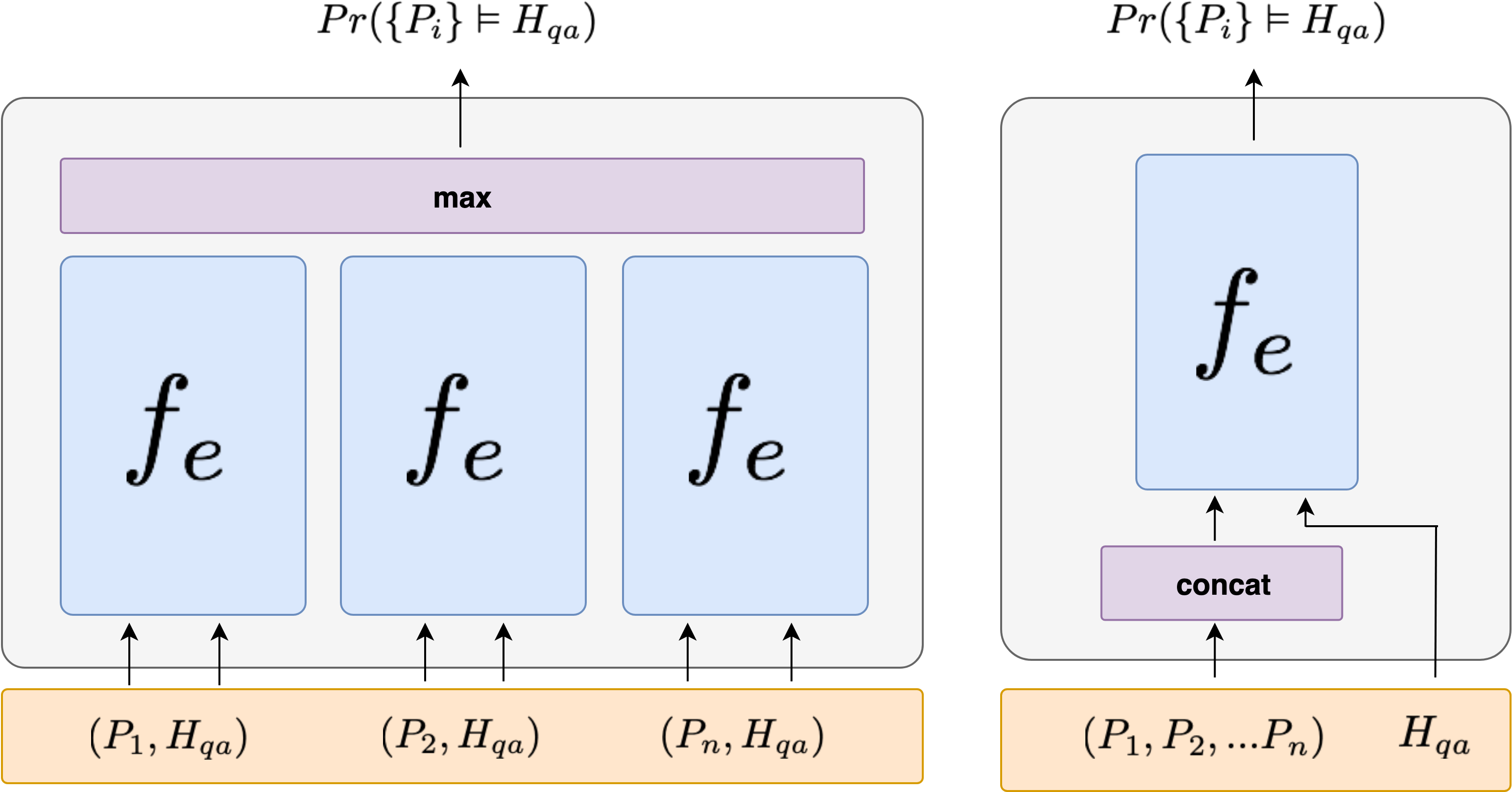}
    \caption{Black Box Applications of Textual Entailment Model for QA: Max and Concat models}
    \label{fig:max-concat}
\end{figure}

\noindent{\bf (i) Aggregate Local Decisions (Max):} Use $f_e$ to check how much each sentence $P_i$ entails $H_{qa}$ on its own, and aggregate these local entailment decisions, for instance, using a max operation.
\begin{equation}
    g_e(\{P_i\}, H_{qa}) = \max_i f_e(P_i, H_{qa}) \label{eqn:maxmodeleqn}
\end{equation}

\noindent{\bf (ii) Concatenate Premises (Concat):} Combine the premise sentences in a sequence to form a single large passage $P$, and use $f_e$ to check whether this passage as a whole entails the hypothesis $H_{qa}$, making a single entailment decision:
\begin{equation}
    g_e(\{P_i\}, H_{qa}) = f_e(P, H_{qa}) \label{eqn:concatmodeleqn}
\end{equation}

Our experiments reveal, however, that neither approach is an effective means of using pre-trained entailment models for QA (see Table~\ref{table:main-results}). For the example in Figure \ref{fig:example}, \textit{Max} model would not be able to consider information from P1 and P3 together. Instead, it will pickup \textit{Silicon Valley} as the answer since P2 is close to $H_s$, \textit{``Facebook was launched in Silicon Valley"}. Similarly, \textit{Concat} would also be muddled by distracting information in P2, which will weaken its confidence in answer \textit{Cambridge}. Therefore, without careful guidance, simple aggregation can easily add distracting information into the premise representation, causing entailment to fail. This motivates the need for new, effective mechanisms for global reasoning over a collection of premises.

\section{Our Approach: \sys}

\begin{figure*}[t!]
    \centering
        \includegraphics[width=\textwidth]{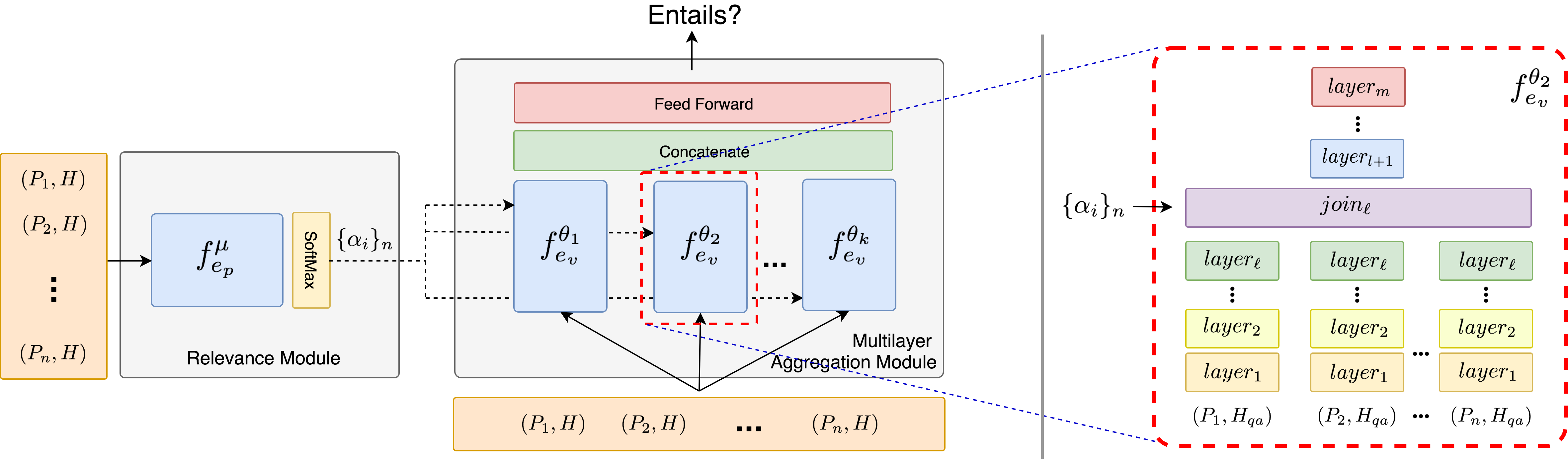}
        \caption{\sys overview: Multee includes two main components, a relevance module, and a multi-layer aggregator module. Both modules use pre-trained entailment functions ($f_{e_p}$ and $f_{e_v}$). $f_{e_p}$ is the full entailment model that gives entailment probability, and $f_{e_v}$ is part of it excluding last projection to logits and softmax. The multi-level aggregator uses multiple copies of entailment function $f_{e_v}$, one for each sub-aggregator performing a \textit{join} at a different layer. Right part of figure zooms in on one such sub-aggregator joining at layer $\ell$.}
        \label{fig:multee-overview-extended}
\end{figure*}

We propose a new entailment based QA model, \sys, with two components: (i) {\bf a sentence relevance model}, which learns to focus on the relevant sentences, and (ii) {\bf a multi-layer aggregator}, which uses an entailment model to obtain multiple layers of question-relevant representations for the premises and then composes them using the sentence-level scores from the relevance model.
Finding relevant sentences is a form of local entailment between each premise and the answer hypothesis, whereas aggregating question-relevant representations is a form of global entailment between all premises and the answer hypothesis. 
This means, we can effectively re-purpose the same pre-trained entailment function $f_e$ for both components. Figure~\ref{fig:multee-overview-extended} shows an architecture that uses multiple copies of $f_e$ to achieve this.

\subsection{Sentence Relevance Model} 
The goal of this module is to identify sentences in the paragraph that are important for the given hypothesis. As shown in Figure~\ref{fig:example}, this helps the global module aggregate relevant content while reducing the chances of picking up distracting information. A sentence is considered important if it contains information that is relevant to answering the question. In other words, the importance of a sentence can be modeled as its entailment probability, i.e., how well the sentence by itself supports the answer hypothesis. We can use a pre-trained entailment model to obtain this. The importance $\alpha_i$ of a sentence $P_i$ can be modeled as:
\begin{align}
\alpha_i &= f_e(P_i, H_{qa})
\end{align}

This can be further improved by modeling the sentence with its surrounding context. This is especially useful for passage-level QA, where the neighboring sentences provide useful context. Given a premise sentence $P_i$, the entailment function $f_e$ computes a single hypothesis-aware representation $x_i$ containing information in the premise that is relevant to entailing the answer hypothesis $H_{qa}$. This is essentially the output of last layer of neural function $f_e$ before projecting it to logits. We denote this part of $f_e$ that outputs last \textit{vector} representation as $f_{e_v}$ and full $f_e$ that gives entailment \textit{probability} as $f_{e_p}$.

We use these hypothesis-aware $x_i$ vectors for each sentence as inputs to a BiLSTM  producing a contextual representation $c_i$ for each premise sentence $P_i$, which is then fed to a feedforward layer that predicts the sentence-level importance as:
\vspace{-0.2cm}
\begin{equation}
    \alpha_i = \softmax(W^T c_i + b)
\end{equation}
\vspace{-0.7cm}

The components for generating $x_i$ are part of the original entailment function $f_e$ and can be pre-trained on the entailment dataset. The BiLSTM to compute $c_i$ and the parameters $W$ and $b$ for computing $\alpha_i$ are not part of the original entailment function and thus can only be trained on the target QA task. We perform this additional \textit{contextualization} only when sentences form contiguous text. Additionally, for datasets such as MultiRC, where the relevant sentences have been marked, we introduce a loss term based on the true relevance label and predicted weights, $\alpha_i$.

\subsection{Multi-level Aggregation}

The goal of this module is to aggregate representations from important sentences in order to make a global entailment decision. There are two key questions to answer: (1) how to combine the sentence-level information into a paragraph-level representation and (2) how to use the sentence relevance weights $\{\alpha_i\}$.

Most entailment models include many layers that transform the input premise and the hypothesis. A typical neural entailment stack includes encoding layers that independently generate contextual representations of the premise and the hypothesis, followed by some cross-attention layer that yields relationships between the premise and hypothesis words, and additional layers that use this cross-attention to generate premise attended representations of the hypothesis and vice versa. The final layers are classification layers which determine entailment based on the representations from the previous layer. Each layer thus generates intermediate representation that captures different type of entailment related information. This presents us with a choice of multiple points for aggregation. 

Figure~\ref{fig:multee-overview-extended} illustrates our approach for aggregating sentence-level representations into a single paragraph level representation. For each premise $P_i$ in the passage, we first process the pair $(P_i, H_{qa})$ through the entailment stack ($f_{e_v}$) resulting in a set of intermediate representations $\{\tilde{X_i}^{\ell}\}$ for each layer $\ell$. We can choose a particular layer $\ell$ to be the aggregation layer. We then compute a weighted combination of the sentence-level outputs at this layer $\{\tilde{X_i}^{\ell}\}$ to produce a passage-level representation $\tilde{Y}^{\ell}$. The weights for the sentences are obtained from the \textit{Sentence Relevance} model. We refer to this as a \textit{join} operation as shown in the Figure \ref{fig:multee-overview-extended}. Layers of the entailment function $f_{e_v}$ that are below the \textit{join} operate at a sentence-level, while layers above the join now operate over paragraph-wise representations. The final layer (i.e. the top most layer) of $f_{e_v}$ thus gives us a vector representation of the entire passage. This type of join can be applied at multiple layers resulting in paragraph vectors that correspond to multiple levels of aggregation. We concatenate these paragraph vectors and pass them through a feed-forward network projecting them down to logits, that can be used to compute the final passage wide entailment probabilities.

\subsubsection{Join Operations}

Given a set of sentence-wise outputs from the lower layer $\{\tilde{X_i}\}$ and the corresponding sentence-relevance weights $\{\alpha_i\}$, the join operation combines them into a single passage-level representation $\tilde{Y}$, which can be directly consumed by the layer above it in the stack. The specifics of the join operation depends on the shape of the outputs from the lower layer, and the shape of the inputs expected by the layer after the join. Here we show four possible join operations, one for each layer.
The ones defined for \textit{Score Layer} and \textit{Embedding Layer} can be reduced to black-box baselines, while we use the other two in \sys.

\noindent \textbf{Score Layer}: The score layer outputs the entailment probabilities  $\{s_i\}$ for each premise to hypothesis independently, which need to be joined to one entailment score. One way to do this is to simply take a weighted maximum of the individual entailment probabilities. So we have $\tilde{X_i} = s_i \quad \forall i$ and $\tilde{Y} = \max_i\big(\alpha_i s_i\big)$.

This reduces to black-box \textit{Max} model (Equation ~\ref{eqn:maxmodeleqn}) when using $\{\alpha_i\} = \mathbf{1}$.

\noindent \textbf{Embedding Layer}: The embedding layer outputs a sequence of embedded vectors of $[\bar{P_i}]$\footnote{We use $[.]$ to denote a sequence and $\bar{.}$ to denote a vector} one sequence for each premise $P_i$ and another sequence of embedded vectors $[\bar{H}_{qa}]$ for the answer hypothesis $H_{qa}$. A join operation in this case scales each embedded vector in a premise by its relevance weight and concatenates them together to form $[\bar{P}]$. $\bar{H}_{qa}$ is passed through unchanged.
\begin{align*}
    \tilde{X_i} & = ([\bar{P_i}], [\bar{H}_{qa}]) \quad \forall i \\
    [\bar{P}]         & = [ \alpha_1[\bar{P_1}] ; \alpha_2[\bar{P_2}] ; \dots ; \alpha_n[\bar{P_n}]] \\
    \tilde{Y}   & = \big( [\bar{P}], [\bar{H}_{qa}] \big)
\end{align*}

For non-contextual word embeddings, this reduces to {\em Concat Premises} (Eq.~\ref{eqn:concatmodeleqn}) when $\{\alpha_i\} = \mathbf{1}$.

\noindent \textbf{Final Layer (FL)}:
The final layer in the entailment stack usually outputs a single vector $\bar{h}$ which is then used in a linear layer and softmax to produce label probabilities. The join operation here is a weighted sum of the premise-level vectors. So we have $\tilde{X_i} = \bar{h_i} \quad \forall i$ and $\tilde{Y} = \sum_i \alpha_i \bar{h_i}$.

This is similar to a standard attention mechanism, where attended representation is computed by summing the scaled representations. However, such scaled addition is not possible when the outputs from lower layers are not of the same shapes, as in the following case.

\noindent \textbf{Cross Attention Layer (CA)}:
Cross-attention is a standard component of many entailment and reading comprehension models. This layer produces three outputs: 
(i) For each premise $P_i$, we get a hypothesis to premise cross attention matrix $M^{hp_i}$ with shape (h $\times$ $p_i$), where $h$ is the number of hypothesis tokens, and $p_i$ is the number of tokens in premise $P_i$; (ii) for each premise $P_i$, we get a sequence of vectors $[\bar{P_i}]$ that corresponds to the token sequence of the premise $P_i$; and (iii) for the hypothesis, we get a single sequence of vectors $[\bar{H}_{qa}]$ that corresponds to its token sequence. $M^{hp_i}$ attention matrix was generated by cross attention from $[\bar{H}_{qa}]$ to $[\bar{P_i}]$.

The join operation in this layer produces a cross attention matrix that spans the entire passage, i.e., has shape ($h \times p$), where $p$ is the total number of tokens across all premises. The operation first scales the cross-attention matrices by the sentence-relevance weights $\{\alpha_i\}$ in order to ``tone down'' the influence of distracting/irrelevant sentences, and then re-normalizes the final matrix:
\begin{align*}
    \tilde{X_i} & = (M^{hp_i}, [\bar{P}_i], [\bar{H}_{qa}]) \quad \forall i \\
    M^{hp} &= \begin{bmatrix} \alpha_i M^{hp_1} ; & \dots ; & \alpha_i M^{hp_n} \end{bmatrix} \\
    M^{hp}_{ij} &= \frac{M^{hp}_{ij}}{\sum_k{M^{hp}_{ik}}} \\
    [\bar{P}] &= \begin{bmatrix} [\bar{P}_1]; & [\bar{P}_2] ; ... ; [\bar{P}_n] \end{bmatrix}\\
    \tilde{Y} &= (M^{hp}, \bar{P}, \bar{H}_{qa})
\end{align*}
where $M^{hp}_{ij}$ is $i^{th}$ row and $j^{th}$ column of $M^{hp}$. 

\sys's multi-layer aggregator module uses join operations at two levels: \textbf{Cross Attention Layer (CA)} and \textbf{Final Layer (FL)}.
The two corresponding aggregators share parameters up till the lower of the two join layers (CA in this case), where they both operate at the sentence level. Above this layer, one aggregator switches to operating at the paragraph level, where it has its own, unshared parameters. In general, if \sys were to aggregate at layers $\ell_{i1}, \ell_{i2}, \dots, \ell_{ik}$, then the aggregators with joins at layers $\ell$ and $\ell'$ respectively could share parameters at layers $1, \ldots, \min \{\ell, \ell'\}$.

\subsection{Implementation Details}

\sys\ uses the ESIM stack as the entailment function pre-trained on SNLI and MultiNLI for both the relevance module and for the multi-layer aggregator module. It uses aggregation at two-levels, one at the cross-attention level (CA) and one at the final layer (FL). All uses of the entailment function in \sys are initialized with the same pre-trained entailment model weights. The embedding layer and the BiLSTM layer process paragraph-level contexts but processing at higher layers are done either at premise level or paragraph-level depending on where the join operation is performed.

\section{Experiments}

\noindent{\bf Datasets:} We evaluate \sys\ on two datasets, OpenBookQA~\cite{mihaylov2018can} and MultiRC~\cite{MultiRC2018}, both of which are specifically designed to test reasoning over multiple sentences.  MultiRC is paragraph-based multiple-choice QA dataset derived from varying topics where the questions are answerable based on information from the paragraph. In MultiRC, each question can have more than one correct answer choice, and so it can be viewed as a binary classification task (one prediction per answer choice), with 4,848 / 4,583 examples in Dev/Test sets. OpenBookQA, on the other hand, has multiple-choice science questions with exactly one correct answer choice and no associated paragraph. As a result, this dataset requires the relevant facts to be retrieved from auxiliary resources including the open book of facts released with the paper and other sources such as WordNet~\cite{miller1995wordnet} and ConceptNet~\cite{L12-1639}. It contains 500 questions in the Dev and Test sets.

\newcommand*{\imgIntable}{\raisebox{-2.0px}{\includegraphics[height=8px,width=8px]{images/MUlTeE-Icon.png}}}

\begin{table*}[t!]
    \centering
    \ra{1.3}
    \setlength\tabcolsep{7pt}
    \small
    \begin{tabular}{llccccccc}\toprule
        & & \multicolumn{2}{c}{OpenBookQA} & \multicolumn{2}{c}{MultiRC} \\
         \cmidrule{3-4} \cmidrule{5-6}
        & & \multicolumn{2}{c}{Accuracy} & \multicolumn{2}{c}{F1a $|$ F1m $|$ EM} \\
        & & Dev & Test & Dev & Test \\
        \midrule

        Entailment           & \imgIntable \sys  & \textbf{56.2}   & 54.8 & 69.6 $|$ 73.0 $|$ \textbf{22.8} & \textbf{70.4} $|$ \textbf{73.8} $|$ \textbf{24.5}\\
        Based Models         & Concatenate Premises  & 47.2 & 47.6 & 68.3 $|$ 71.3 $|$ 17.9   & 68.5 $|$ 72.5 $|$ 18.0  \\
        with ELMo               & Max of Local Decisions  & 47.8 & 45.2 & 66.5 $|$ 69.3 $|$ 16.3 & 66.70 $|$ 70.4 $|$ 19.4 \\
        \midrule
        Entailment           & \imgIntable \sys       & 54.6 & \textbf{55.8} & 68.3 $|$ 71.7 $|$ 16.4 & 69.9 $|$ 73.6 $|$ 19.0    \\ 
        Based Models         & Concatenate Premises  & 47.4 & 42.6 & 66.9 $|$ 70.7 $|$ 14.8 & 68.7 $|$ 72.4 $|$ 16.3  \\
        with GloVe           & Max of Local Decisions & 44.4 & 47.6 & 66.8 $|$ 70.3 $|$ 16.9 & 67.7 $|$ 71.1 $|$ 18.2 \\
        \midrule
        Previously           & LR \cite{MultiRC2018}  & --- & --- & 63.7 $|$ 66.5 $|$ 11.8 & 63.5 $|$ 66.9 $|$ 12.8\\
        Published            & IR \cite{MultiRC2018}  & --- & --- & 60.0 $|$ 64.3 $|$ \phantom{0.00} & 54.8 $|$ 53.9 $|$ \phantom{0.00} \\
        Results              & QM + ELMo \cite{mihaylov2018can}  & 54.6 & 50.2 & --- & --- \\
                             & ESIM + ELMo \cite{mihaylov2018can} & 53.9 & 48.9 & --- & --- \\
                             & KER \cite{mihaylov2018can}& 55.6 & 51.4 & --- & --- \\
        \midrule
        Large                & OFT \cite{sun2018improving}  & --- & 52.0 & 67.2 $|$ 69.3 $|$ 15.2& --- \\
        Transformer          & OFT (ensemble) \cite{sun2018improving}   & --- & \phantom{$^*$}52.8$^*$ & \phantom{$^*$}67.7 $|$ 70.3 $|$ 16.5$^*$& --- \\
        Models               & RS \cite{sun2018improving}  & --- & \phantom{$^*$}55.2$^*$ & \phantom{$^*$}69.2 $|$ 71.5 $|$ 22.6$^*$& --- \\
                             & RS (ensemble) \cite{sun2018improving}  & --- & \phantom{$^*$}\textbf{55.8}$^*$ & \phantom{$^*$}\textbf{70.5} $|$ \textbf{73.1} $|$ 21.8$^*$& --- \\
        \bottomrule
    \end{tabular}
    \caption{Comparison of \sys with other systems. Starred (*) results are based on contemporaneous system. Results marked (---) are not available. RS is Reading Strategies, KER is Knowledge Enhanced Reader, OFT is OpenAI FineTuned Transformer.}
    \label{table:main-results}
\end{table*}

\noindent \textbf{Preprocessing:} For each question and answer choice, we create an answer hypothesis statement using a modified version of the script used in SciTail~\cite{khot2018scitail} construction. We wrote a handful of rules to better convert the question and answer to a hypothesis. We also mark the span of answer in the hypothesis with special begin and end tokens, \texttt{@@@answer} and \texttt{answer@@@}  respectively\footnote{Answer span marking gave substantial gains for all entailment based models including the baselines.}. For MultiRC, we also apply an off-the-shelf coreference resolution model\footnote{https://github.com/huggingface/neuralcoref} and replace the mentions when they resolve to pronouns occurring in a different sentence\footnote{It is hard to learn co-reference, as these target datasets are too small to learn this in an end-to-end fashion.}. For OpenBookQA, we use the exact same retrieval as released by the authors of OpenBookQA\footnote{https://github.com/allenai/OpenBookQA} and use the OpenBook and WordNet as the knowledge source with top 5 sentences retrieved per query.

\noindent \textbf{Training \sys:} For OpenBookQA we use cross entropy loss for labels  corresponding  to  4  answer  choices.    For MultiRC,  we  use  binary  cross  entropy loss  for each answer-choice separately since in MultiRC each question can have more than one correct answer choice.  The entailment components are pre-trained on sentence-level entailment tasks and then fine-tuned as part of end-to-end QA training. The MultiRC dataset includes sentence-level relevance labels. We supervise the Sentence Relevance module with a binary cross entropy loss for predicting these relevance labels when available. We used PyTorch \cite{paszke2017automatic} and AllenNLP to implement our models and ran them on Beaker\footnote{\url{https://beaker.org/}}.  For pre-training we use the same hyper-parameters of ESIM\cite{chen2016enhanced} as available in implementation of AllenNLP \cite{Gardner2017AllenNLP} and fine-tune the model parameters. We do not perform any hyper-parameter tuning for any of our models. We fine-tune all layers in ESIM except for the embedding layer.

\noindent{\bf Models Compared:} We experiment with Glove \cite{pennington2014glove} and ELMo \cite{Peters:2018} embeddings for \sys and compare with following three types of systems:

\noindent \textbf{(A) Baselines using entailment as a black-box}
We use the pre-trained entailment model as a black-box in two ways: concatenate premises (\textit{Concat}) and aggregate sentence level decisions with a max operation (\textit{Max}). Both models were also pre-trained on SNLI and MultiNLI datasets and fine-tuned on the target QA datasets with same pre-processing.

\noindent \textbf{(B) Previously published results:}
For MultiRC, there are two published baselines: IR (Information Retrieval) and LR (Logistic Regression). These simple models turn out to be strong baselines on this relatively smaller sized dataset.
For OpenBookQA, we report published baselines from ~\cite{mihaylov2018can}: Question Match with ELMo (QM + ELMo),  Question to Answer ESIM with ELMo (ESIM + ELMo) and their best result with the Knowledge Enhanced Reader  (KER). 

\noindent \textbf{(C) Large Transformer based models:}
We compare with OpenAI-Transformer (OFT), pre-trained on large-scale language modeling task and fine-tuned on respective datasets. A contemporaneous work,\footnote{Published on arXiv on Oct 31, 2018~\cite{sun2018improving}.} which published these transformer results, also fine-tuned this transformer further on a large scale reading comprehension dataset, RACE~\cite{lai2017race}, before fine-tuning on the target QA datasets with their method, \textit{Reading Strategies}.

\subsection{Results}

Table~\ref{table:main-results} summarizes the performance of all models. \sys outperforms the black-box entailment baselines (Concat and Max) that were pre-trained on the same data, previously published baselines, OpenAI transformer models. We note that the 95\% confidence intervals around baseline accuracy for OpenBookQA and MultiRC are 4.3\% and 1.3\%, respectively.

On OpenBookQA test set, \sys with GloVe outperforms ensemble version of OpenAI transformer by 3.0 points in accuracy. It also outperforms single model version of Reading Strategies system and is comparable to their ensemble version. On MultiRC dev set, \sys with ELMo outperforms ensemble version of OpenAI transformer by 1.9 points in F1a, 2.7 in F1m and 6.3 in EM. It also outperforms single model version of Reading Strategies system and is comparable to their ensemble version. Recall that the Reading Strategies results are reported with an additional fine-tuning on another larger QA dataset, RACE~\cite{lai2017race} aside from the target QA datasets we use here.

While ELMo contextual embeddings helped in MultiRC, it did not help OpenBookQA. We believe this is in part due to the mismatch between our ELMo training setup where all sentences are treated as a single sequence, which, while true in MultiRC, is not the case in OpenBookQA.

In general, gains from \sys are more prominent in OpenBookQA than in MultiRC. We hypothesize that a key contributor to this difference is distraction being a lesser challenge in MultiRC, where premise sentences come from a single paragraph whose other sentences are often irrelevant and rarely distract towards incorrect answers. OpenBookQA has a noisier set of sentences, since an equal number of sentences is retrieved for the correct and each incorrect answer choice.

\subsection{Ablations}

\noindent{\bf Relevance Model Ablation.} 
Table~\ref{table:ablations-relevance-model} shows the utility of the relevance module. We use the same setting as the full model (aggregation at Cross Attention (CA) and the Final Layer (FL)). As shown in the table, using the relevance module weights (\cmark $\alpha_i$) leads to improved accuracy on both datasets (substantially so in OpenBookQA) as compared to ignoring the module, i.e., setting all weights to 1 (\xmark $\alpha_i$). In MultiRC, we show that the additional supervision for the relevance module leads to even further improvements in score.

\begin{table}[t!]
    \centering
    \ra{1.3}
    \small
    \begin{tabular}{rcc}\toprule
        & OpenBookQA & MultiRC \\
         \cmidrule{2-2} \cmidrule{3-3}
        & Accuracy & F1a $|$ F1m \\ \midrule
        \xmark $\alpha_i$ & 50.6 & 67.3 $|$ 70.3 \\
        \cmark $\alpha_i$ & \textbf{55.8} & 67.4 $|$ 71.0 \\
        \cmark $\alpha_i$ + supervise  & --- & \textbf{68.3} $|$ \textbf{71.7} \\
        \bottomrule
    \end{tabular}
    \caption{Relevance Model Ablation of \sys. \xmark $\alpha_i$: without relevance weights, \cmark $\alpha_i$: with relevance weights respectively, \cmark $\alpha_i$ + supervise: with supervised relevance weights. Test results on OpenBookQA and Dev results on MultiRC.}
    \label{table:ablations-relevance-model}
\end{table}
\begin{table}[t!]
    \centering
    \ra{1.3}
    \small
    \begin{tabular}{rccc}\toprule
        \multirow{2}{*}{Aggregator} & OpenBookQA & MultiRC \\
         \cmidrule{2-2} \cmidrule{3-3}
        & Accuracy & F1a $|$ F1m \\ \midrule
        Cross Attention (CA) & 45.8 & 67.2 $|$ 71.1 \\
        Final Layer (FL) & 51.0 & 68.3 $|$ 71.5 \\
        CA +FL & \textbf{55.8} & \textbf{68.3} $|$ \textbf{71.7} \\
        \bottomrule
    \end{tabular}
    \caption{Aggregator Level Ablation of \sys. On MultiRC, \sys uses relevance supervision but not on OpenBookQA because of unavailibility. Test results on OpenBookQA and Dev results on MultiRC.}
    \label{table:ablations-aggregator}
\end{table}

\noindent{\bf Multi-Level Aggregator Ablation.} \sys performs aggregation at two levels: Cross Attention Layer (CA) and Final Layer (FL). We denote this by CA+FL. To show that multi-level aggregation is better than individual aggregations, we train models with aggregation at only FL and at only CA. Table~\ref{table:ablations-aggregator} shows that multi-layer aggregation is better than CA or FL alone on both the  datasets.  

\subsection{Effect of Pre-training}

One of the benefits of using entailment based components in a QA model is that we can pre-train them on large scale entailment datasets and fine-tune them as part of the QA model. Table ~\ref{table:pre-training} shows that such pre-training is valuable. The model trained from scratch is substantially worse in the case of OpenBookQA, highlighting the benefits of our entailment-based QA model.

\sys benefits come from two sources: (i) Re-purposing of entailment function for multi-sentence question answering, and (ii) transferring from a large-scale entailment task. In the case of OpenBookQA, both are helpful. For MultiRC, only the first is a significant contributor. Table \ref{table:baseline-pre-training} shows that re-purposing was a bigger factor for MultiRC, since Max and Concat models do not work well when trained from scratch.

\begin{table}[t!]
    \centering
    \ra{1.3}
    \small
    \begin{tabular}{@{}rcc}\toprule
        & OpenBookQA & MultiRC \\
        & Accuracy & F1a $|$ F1m \\ \midrule
        Snli + MultiNli & \textbf{55.8} & \textbf{69.9} $|$ \textbf{73.6} \\
        Snli            & 50.4 & 69.3 $|$ 73.3 \\
        Scratch         & 42.2 & 68.3 $|$ 72.6 \\
        \bottomrule
    \end{tabular}
    \caption{Effect (on test data) of pre-training the entailment model used in \sys.}
    \label{table:pre-training}
\end{table}

\begin{table}[ht]
    \centering
    \ra{1.3}
    \small
    \begin{tabular}{@{}rrcc}\toprule
        & & OpenBookQa & MultiRc \\
        & & Accuracy & F1a $|$ F1m \\ \midrule
        \multirow{2}{*}{Max} & Snli + MultiNli & \textbf{47.6} & \textbf{66.8} $|$ \textbf{70.3} \\
                             & Scratch         & 32.4 & 42.8 $|$ 44.0 \\
        \midrule
        \multirow{2}{*}{Concat} & Snli + MultiNli & \textbf{42.6} & \textbf{66.9} $|$ \textbf{70.7} \\
                                & Scratch         & 35.8 & 51.3 $|$ 50.4 \\
        \bottomrule
    \end{tabular}
    \caption{Pre-training ablations of black-box entailment baselines for OpenBookQA (test) and MultiRC (dev).}
    \label{table:baseline-pre-training}
\end{table}

\begin{table}[t!]
    \centering
    \ra{1.3}
    \small
    \begin{tabular}{@{}ccc}\toprule
                & F1a precision & F1a recall \\ \midrule
        IR Sum Loss  & \textbf{59.5} & 68.5 \\
        BCE Loss & 58.0 & \textbf{83.2} \\
        \bottomrule
    \end{tabular}
    \caption{F1a precision and recall on MultiRC Dev with 2 kinds of relevance losses. IR Sum is the sum of attention probability mass on irrelevant sentences. BCE is Binary Cross Entropy loss.}
    \label{table:coverage-loss}
\end{table}

\begin{figure*}[t!]
    \centering
    \begin{subfigure}{.48\textwidth}
        \centering
        \includegraphics[width=\linewidth]{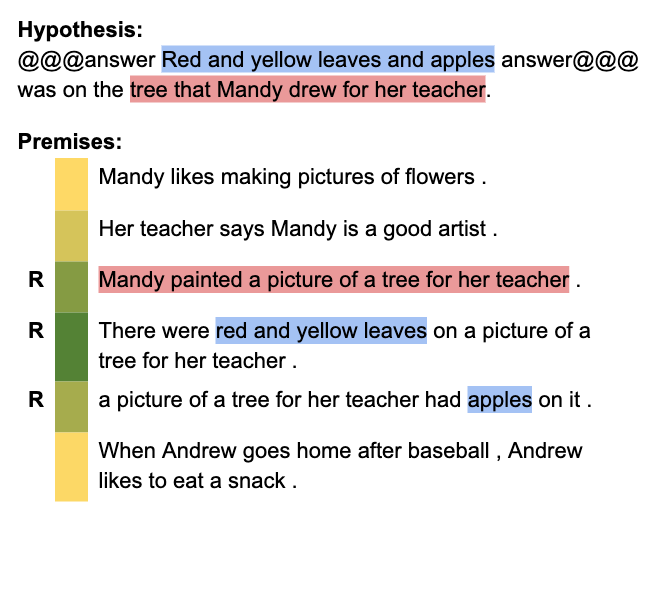}
        \vspace{-2em}
        \caption{Positive Example}
        \label{fig:pos_ex}
    \end{subfigure}%
    \hfill
    \begin{subfigure}{.48\textwidth}
        \centering
        \includegraphics[width=\linewidth]{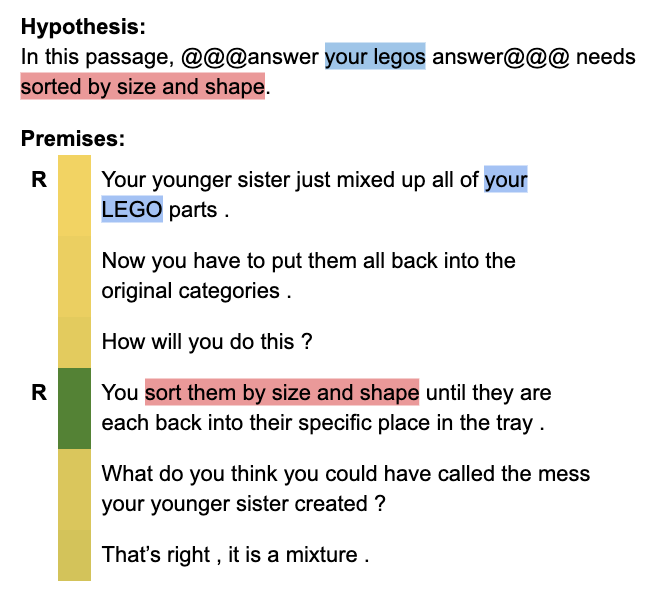}
        \vspace{-2em}
        \caption{Negative Example}
        \label{fig:neg_ex}
    \end{subfigure}
    \caption{Success and failure examples of \sys from MultiRC. \textbf{R}: annotated relevant sentences. Green/yellow: high/low predicted relevance.}
    \label{fig:examples}
\end{figure*}

\section{Analysis}

\begin{figure}
    \centering
    \includegraphics[width=.95\linewidth]{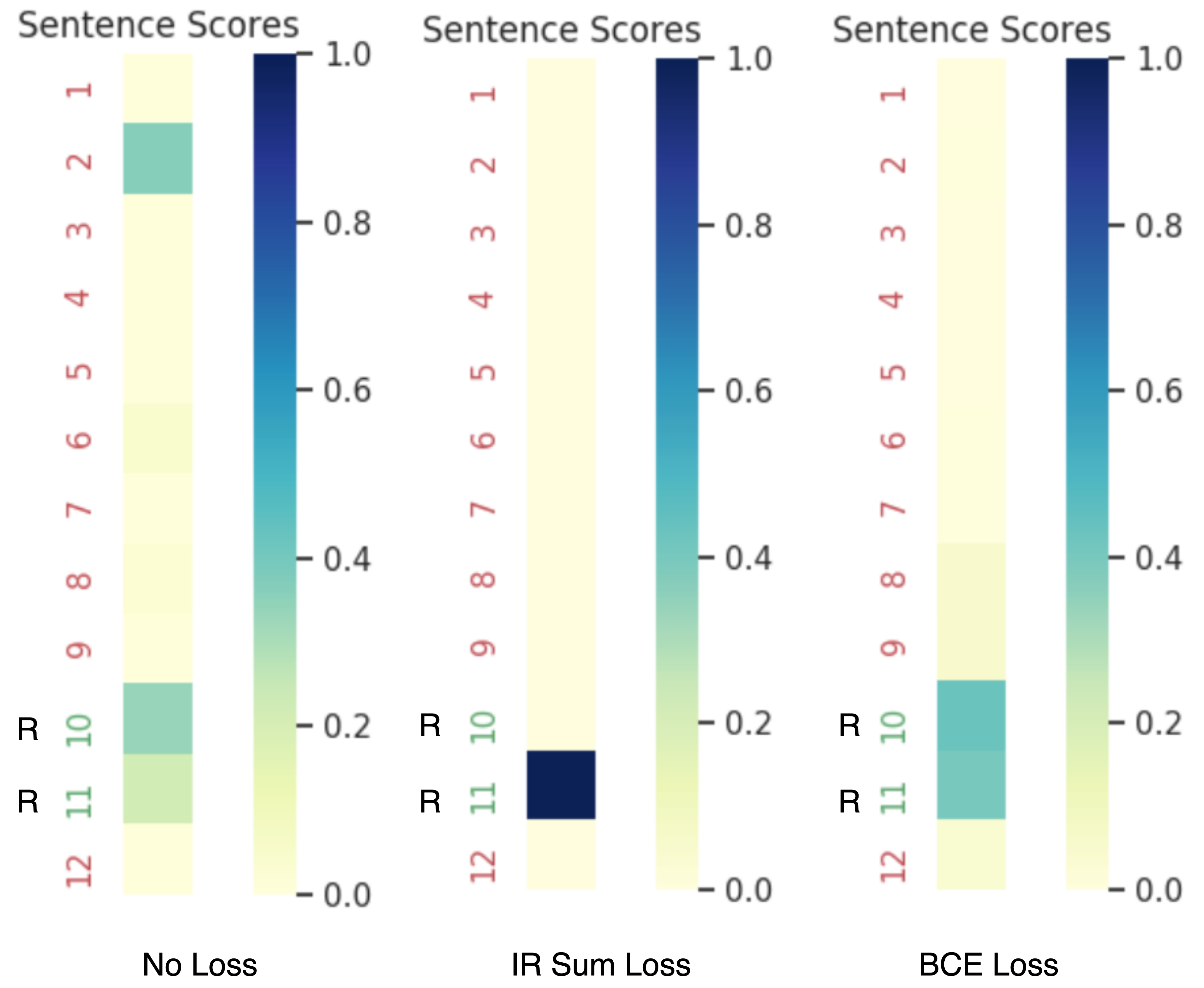}
    \caption{Sentence level attentions for various sentence relevance losses. \textbf{R}: annotated relevant sentences.}
    \label{fig:coverage-loss}
\end{figure}

\noindent{\bf Relevance Loss.} The sentence-level relevance model provides a way to dig deeper into the overall QA model's behavior. When sentence-level supervision is available, as in the case of MultiRC, we can analyze the impact of different auxiliary losses for the relevance module. Table~\ref{table:coverage-loss} shows the QA performance with different relevance losses, and Figure \ref{fig:coverage-loss} shows a visualization of attention scores for a question in MultiRC. Overall, we find that two types of behaviors emerge from different loss functions. For instance, trying to minimize the sum of attention probability mass on irrelevant sentences i.e. $\sum_i \alpha_i (1 - y_i)$, called \emph{IR Sum Loss}, causes the attention scores to become "peaky" i.e, high for one or two sentences, and close to zero for others. This leads to higher precision but at significantly lower recall for the QA system, as it now uses information from fewer but highly relevant sentences. Binary cross entropy loss (BCE) allows the model to attend to more relevant sentences thereby increasing recall without too much drop in precision. 

\noindent{\bf Failure Cases.} As Figure~\ref{fig:coverage-loss} shows, our model with BCE loss tends to distribute the attention, especially to sentences close to the relevant ones. We hypothesize that the model is learning to use the contextualized BiLSTM representations to incorporate information from neighboring sentences, which is useful for this task and for passage understanding in general. For example, more than 60\% of Dev questions in MultiRC have at least one adjacent relevant sentence pair. Figure~\ref{fig:pos_ex} illustrates this behavior.

On the other hand, if the relevant sentences are far apart, the model finds it difficult to handle such long-range cross sentence dependencies in its contextualized representations. As a result, it ends up focusing attention on the most relevant sentence, missing out on other relevant sentences (Figure~\ref{fig:neg_ex}). When these unattended but relevant sentences contain the answer, the model fails.

\section{Related Work}

Entailment systems have been applied to question-answering before but have only had limited success~\cite{harabagiu2006methods,sacaleanu2008entailment,Clark2012AnEA} in part because of the small size of the early entailment datasets~\cite{dagan2006pascal,dagan2013recognizing}. Recent large scale entailment datasets such as SNLI~\cite{bowman2015large} and MultiNLI~\cite{williams2017broad} have led to many new powerful neural entailment models that are not only more effective, but also produce better representations of sentences~\cite{conneau-EtAl:2017:EMNLP2017}. Models such as Decomposable Attention~\cite{parikh2016decomposable} and ESIM~\cite{chen2016enhanced}, on the other hand, find alignments between the hypothesis and premise words through cross-attention. However, these improvements in entailment models have not yet translated to improvements in end tasks such as question answering. 

SciTail~\cite{khot2018scitail} was created from a science QA task to push for models with a direct impact on QA. Entailment models trained on this dataset show minor improvements on the Aristo Reasoning Challenge~\cite{clark2018think, Musa2018AnsweringSE}. However, these QA systems make independent predictions and can not combine information from multiple supporting sentences. 

Combining information from multiple sentences is a key problem in language understanding. Recent Reading comprehension datasets~\cite{welbl2018qangaroo,MultiRC2018,yang2018hotpotqa,mihaylov2018can}  explicitly evaluate a system's ability to perform such reasoning through questions that need information from multiple sentences in a passage. Most approaches on these tasks perform simple attention-based aggregation~\cite{mihaylov2018can,mhqa_grn,gcn_entity} and do not exploit the entailment models trained on large scale datasets.

\section{Conclusions}
 
Using entailment for question answering has seen limited success. Neural entailment models are designed and trained on tasks defined over sentence pairs, whereas QA often requires reasoning over longer texts spanning multiple sentences. We propose \sys, a novel QA model that addresses this mismatch. It uses an existing entailment model to both focus on relevant sentences and aggregate information from these sentences. Results on two challenging QA datasets, as well as our ablation study, indicate that entailment based QA can achieve state-of-the-art performance and is a promising direction for further research.

\subsection*{Acknowledgements}

This work is supported in part by the National Science Foundation under Grant IIS-1815358. The computations on beaker.org were supported in part by credits from Google Cloud.

\bibliographystyle{acl_natbib}
\bibliography{Multee}

\end{document}